\documentclass[letterpaper, 10 pt, conference]{ieeeconf}  

\IEEEoverridecommandlockouts                              

\usepackage{graphicx}
\usepackage{float}
\usepackage{amsmath}
\usepackage[section]{placeins}
\usepackage{lipsum} 
\usepackage{environ} 
\usepackage{amsmath}
\usepackage{tabularx}
\usepackage{adjustbox}
\usepackage[dvipsnames]{xcolor}
\usepackage{comment} 
\usepackage{setspace}
\usepackage{fixltx2e}
\usepackage{balance}

\title{\LARGE \bf
A Complementarity-Based Switch-Fuse System for Improved Visual Place Recognition }

\author{Maria Waheed$^{1}$, Sania Waheed$^{2}$,  Michael Milford$^{3}$,  Klaus McDonald-Maier$^{1}$ and Shoaib Ehsan$^{1,4}$
\thanks{*This research was supported by the UK Engineering and Physical Sciences Research Council (EPSRC) through grants EP/R02572X/1,  EP/P017487/1 and EP/V000462/1. This research was also partially supported by funding from ARC Laureate Fellowship FL210100156 to MM and the QUT Centre for Robotics. \textit{(Corresponding author: Maria Waheed)}}
\thanks{$^{1}$M. Waheed, K. McDonald-Maier and S. Ehsan are with the School of Computer Science and Electronic Engineering, University of Essex, Colchester CO4 3SQ, United Kingdom.
        {\tt\small (e-mail: mw20987@essex.ac.uk; kdm@essex.ac.uk; sehsan@essex.ac.uk;  )}}%
\thanks{$^{2}$S. Waheed is with the Kim Jaechul Graduate School of AI, Korea Advanced Institute of Science and Technology, 207-42 Cheongnyangni-dong, Dongdaemun-gu, Seoul
         {\tt\small (e-mail: saniawaheed@kaist.ac.kr )}}%
\thanks{$^{3}$M. Milford is with the School of Electrical Engineering and Computer
Science, Queensland University of Technology, Brisbane, QLD 4000, Australia
        {\tt\small (e-mail: michael.milford@qut.edu.au)}}%
\thanks{$^{4}$Shoaib Ehsan is also with the School of Electronics and Computer Science, University of Southampton, Southampton, SO17 1BJ
        {\tt\small (e-mail: s.ehsan@soton.ac.uk)}}        
}

\begin{document}

\maketitle
\thispagestyle{empty}
\pagestyle{empty}

\begin{abstract}

Recently several fusion and switching based approaches have been presented to solve the problem of Visual Place Recognition. In spite of these systems demonstrating significant boost in VPR performance they each have their own set of limitations. The multi-process fusion systems usually involve employing brute force and running all available VPR techniques simultaneously while the switching method attempts to negate this practise by only selecting the best suited VPR technique for given query image. But switching does fail at times when no available suitable technique can be identified. An innovative solution would be an amalgamation of the two otherwise discrete approaches to combine their competitive advantages while negating their shortcomings. The proposed, Switch-Fuse system, is an interesting way to combine both the robustness of switching VPR techniques based on complementarity and the force of fusing the carefully selected techniques to significantly improve performance. Our system holds a structure superior to the basic fusion methods as instead of simply fusing all or any random techniques, it is structured to first select the best possible VPR techniques for fusion, according to the query image. The system combines two significant processes, switching and fusing VPR techniques, which together as a hybrid model substantially improve performance on all major VPR data sets illustrated using PR curves.

\end{abstract}

\section{INTRODUCTION}

Visual Place Recognition is a rapidly growing field of research with interest from both computer vision and robotics communities [1]. The task of recognizing a previously visited location using visual information is the most basic explanation for VPR. With ongoing researches and breakthroughs being made every year, it is still however an open problem. With several different types of variation including severe viewpoint, illumination and seasonal changes [2]-[9], all factors that add to the complex nature of the VPR problem. There is a large group of existing VPR techniques some of which have excellent performance however none of the techniques are robust to all types of variations in an environment. Leaving us a with a pool of strong VPR techniques but each with its own set of perks and downsides. 

\begin{figure} [tb]
    \centering
    \vspace*{1mm}
    \includegraphics[width=1\columnwidth]{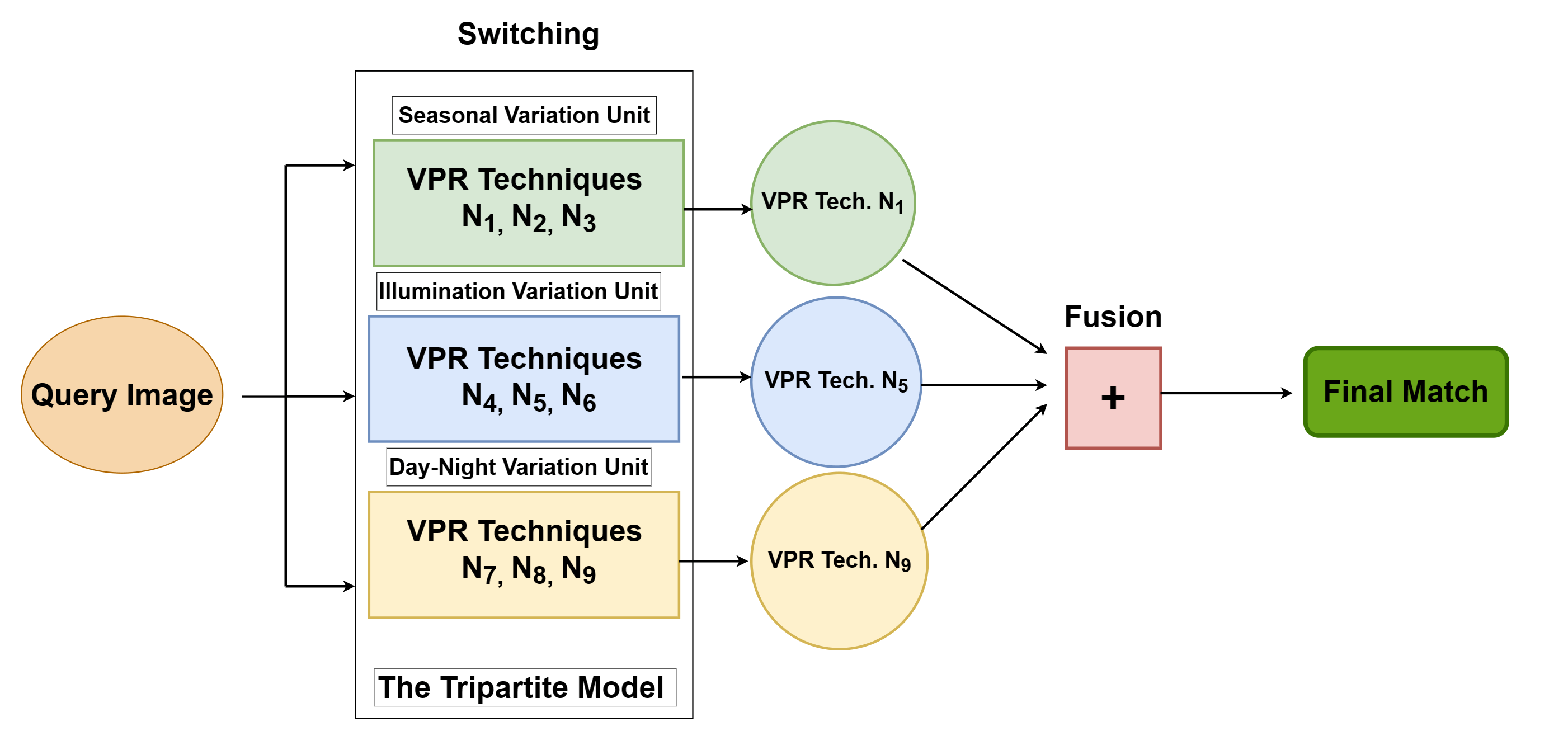}
    \caption{Our \textbf{Switch-Fuse System} is a \textit{\textbf{tripartite model}} consisting of three well curated tiers based on complementarity and environmental variations with each \textit{\textbf{unit}} of the model determining one VPR technique with the highest probability of correctly matching the query image. The determined VPR techniques then undergo \textit{\textbf{fusion}} which involves using their feature vectors to calculate the combined similarity vectors to determine the correct match thus improving performance.}
    \label{fig:my_label}
\end{figure} 

Hence recent research endeavours have shifted their focus from developing an entirely new VPR technique to utilizing the existing techniques to achieve their maximum potential. Very interesting work in this regards has been presented by [10],[11] that introduces the concept of multi-process fusion between different VPR techniques. Along side with a detailed analysis by [12] on the complementarity of the existing VPR techniques providing information on different complementary VPR technique pairs. The analysis is based on the fact observed through empirical data that different techniques often have complementary performances and randomly selecting different VPR techniques for fusion can actually have a detrimental effect on performance and computation in some cases [10],[11]. For example fusing techniques that have a redundant performance will most likely waste the fusion effort and will be unable to improve performance. Hence having this knowledge of which techniques complement each other is the key to developing more efficient fusion systems. Recently, another innovative approach based on combining abilities of different VPR techniques by merely switching to a better technique based on complementarity, SwicthHit [13], has been presented and it depicts promising results. Although [10]-[13] are all substantial research efforts attempting to improve performance by creating a superior VPR system there are still shortcomings to address and a long way before an actual super system can be created.

Our proposed system, Switch-Fuse, is based on observing that both fusion and switching methods have the potential to improve performance. Switching takes precedence over fusion in some cases as it merely shifts to the better performing algorithm [13] and ends up selecting just one rather than running all or multiple techniques at once to combine their results. But simultaneously switching does fail at times when all available techniques have a probability of correct match that is below threshold, a major shortcoming faced by [13]. In such a case [13] selects the technique with the highest probability to use which is not optimal. However, introducing fusion to the scenario using Switch-Fuse mitigates this problem. 

Based on these observations, we are proposing a system with three units carefully designed to incorporate a variety of VPR techniques, categorized on the basis of their performance for different types of variations. However, the system is not restricted to a variation categorization and can be employed for any other type of classification of different VPR techniques. The switching component allows for the selection of the best suited techniques for fusion to ensure improved performance. Such an approach saves us from having to use brute force and fuse all or some random techniques together, rather it intelligently selects from the pool of techniques in different units using a Bayes theorem inspired framework and fuses the final selected and most suitable techniques only. We present our results in form of performance comparison using PR curves and show the improvement in accuracy compared to SwitchHit [13] and the Multi-Process Fusion system [10],[11] and as well as other individual high performing VPR techniques. 

The rest of this paper is organized as follows. Section II provides an overview of related work. Section III presents the methodology explaining the mechanisms of the Switch-Fuse system. Section IV describes the experimental setup. The results based on the proposed system are presented in Section V and finally, conclusions are given in section VI.

\section{Related Work }
This section provides an overview of the related work in the field of visual place recognition. The earliest literature for VPR is dominated by local and global feature descriptors such as  Scale-Invariant Feature Transformation (SIFT) [14] and Speed-Up Robust Features (SURF) [15] and these descriptors have been used for the VPR problem by [21], [22], [27], and [36]. Additionally many other variants for SURF such as the Whole-Image SURF (WI-SURF) [18] have also been introduced and the different hand-crafted global image descriptors, such as GIST [17] and the bag-of-visual-words [16] approach. These methods, for a long time, have been widely used and presented impressive results. While CoHOG [35] is great example of proposed handcrafted feature descriptor-based technique, which uses image entropy to identify salient regions in an image and then assigns those regions with a HOG [30][34] descriptor. However, more recent times the VPR task has shifted more towards employing Convolutional Neural Networks (CNNs) such as NetVLAD [19] and RegionVLAD [33] which  have turned out to be revolutionary in terms of VPR performance and effectiveness even under extreme environmental variations. A commonly used example for Region-of-interest-based VPR techniques is Regions of Maximum Activated Convolutions (R-MAC) [23]. \begin{figure} [tb]
    \centering
    \vspace*{1mm}
    \includegraphics[width=7cm,height=6cm]{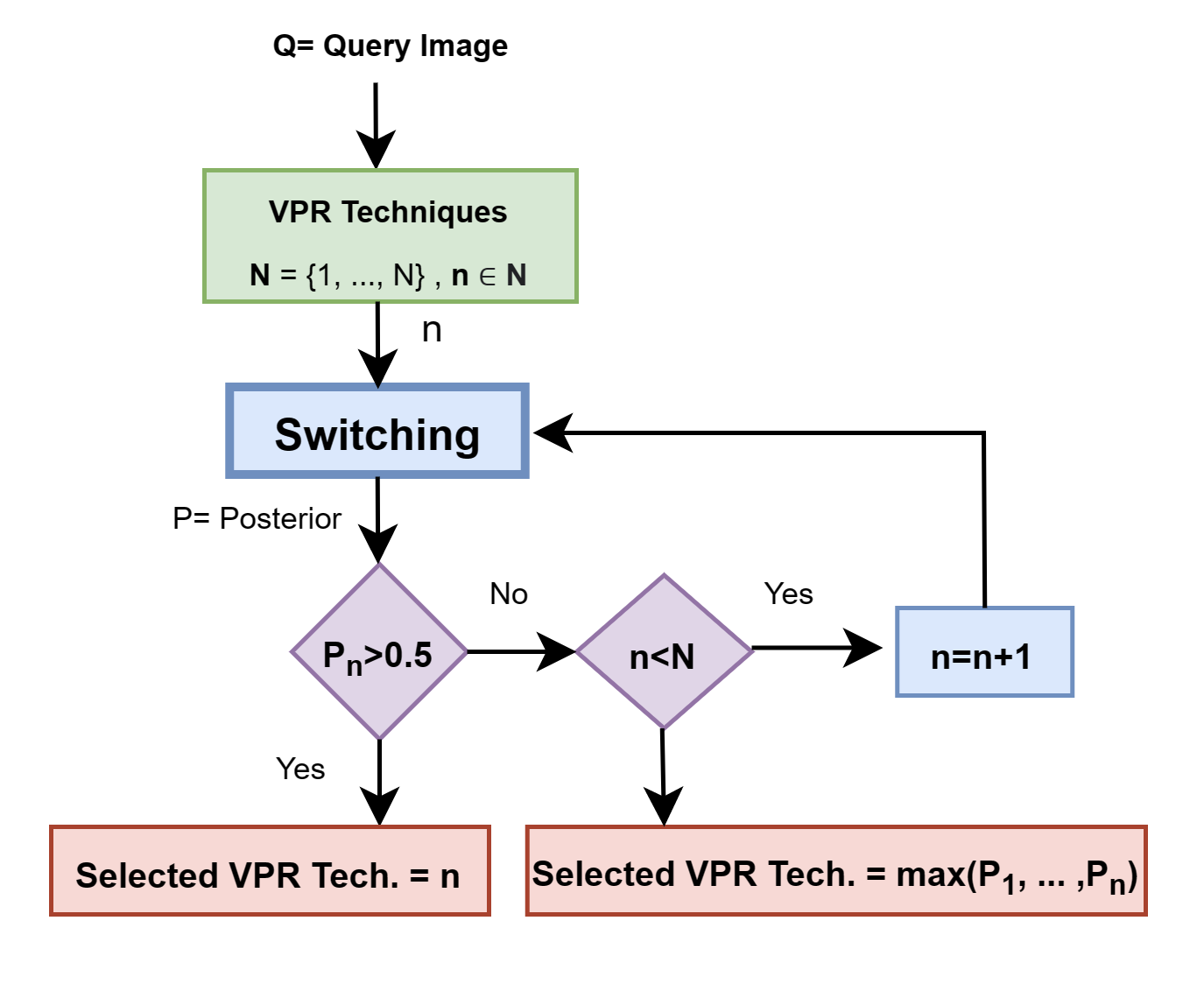}
    \caption{One \textit{\textbf{unit}} of the tripartite model selects the best VPR technique available by calculating the probability of correct match and determining the VPR technique with the highest probability as the final selected VPR Technique.}
    \label{fig:my_label}
\end{figure}

Though an abundant amount of VPR methods exist today it is an ongoing research with room for great improvement. One such idea is the process of multi-fusion within VPR techniques which is primarily inspired by the idea of the multi-sensor approach [32] to solve the place recognition problem. The multi-fusion approach works on the bases of efficiently combining multiple VPR techniques to enhance performance. The authors of [31] combined multiple image processing methods into a merged feature vector using a convex optimization approach to decide the best match from the sequence of images generated. 

Other multi-process fusion systems following this approach are introduced in [10] and [11] using a Hidden Markov Model and a hierarchical system respectively. Furthermore, identifying some limitations of the multi-fusion idea, interesting work was also presented that explores the notion of complementarity between multiple VPR techniques has been introduced by [12]. A McNemar’s test like approach as discussed in [12] is used to test out the level of complementarity between different VPR techniques. The results presented show that employing complementary VPR techniques in a combined VPR setup will result in much more improved results than an otherwise random selection. The most recent endeavour was presented by [13] that presents a system based on complementarity which switches between different VPR techniques to select the best suited technique according to query image. This approach not only significantly improved performance but also was a more efficient way of making use of multiple VPR techniques than using brute force and combining all techniques for improvement.

\begin{table}[]
\centering
\caption{VPR-Bench Datasets [39]} 
\begin{tabular}{|c|c|}
\hline
\textbf{Dataset} & \textbf{Conditional-Variation} \\ \hline
GardensPoint     & Day-Night                      \\ \hline
ESSEX3IN1        & Illumination                   \\ \hline
Cross-Seasons    & Dawn-Dusk                      \\ \hline
Nordland         & Seasonal                       \\ \hline
Corridor         & None                           \\ \hline
Living-room      & Day-Night                      \\ \hline
\end{tabular}
\end{table}

\section{Methodology}

This section presents our Switch-Fuse system that incorporates the robustness of both, a VPR switching system and multi-process fusion. A solo VPR switching system merely focuses on shifting between the available VPR techniques, to the best suited technique according to query image. The system can and does improve performance by applying this intelligent switching approach based on a Bayes theorem inspired framework but it is not without its flaws. As such a system does encounter failure when it is unable to find any suitable technique to select based on its calculated probability scores of correct match. A multi-process fusion system, on the other hand, is different in its operation as it simply fuses the provided VPR techniques through the similarity vectors to get the correct match hence does not encounter the same problems. However, running a purely multi-fusion system requires brute force to run and fuse all available techniques. A preferable approach would be to target the middle ground between these two otherwise discrete methodologies and utilize the strengths of both ideas.

Our system begins with a tripartite model with each component that further consists of multiple VPR techniques for our experimentation. They have been categorized on the basis of their performance on different types of environmental variations as illustrated in Fig. 1. 
The query image is input to all three units of the system, where the probability of correct match is calculated by the primary technique in each unit and switching is conducted to select an alternate technique when required as illustrated in Fig. 2. Finally a single technique is selected by each unit. The three units select one VPR technique, each of which has the highest likelihood of correctly matching the query image. These selected techniques then undergo fusion where we add the normalized distance vectors produced by each of these techniques, as displayed in Fig. 3, to ensure a significant enhancement in performance. 

Our system can be majorly divided into two main steps starting from performing switching for each unit of the model and then the selected VPR techniques undergo fusion to determine the correct match for the query image. 

\subsection{Input Query Image to Each Component of the Tripartite Model}
The first step is the query image provided as an input to each of three parts of the system. The query image then undergoes several steps and calculations for the final selection of a best suited VPR technique from each component. 

Let X be the set of query images in a data set and Z be the set of components of the tripartite model performing switching. 

\begin{equation}
    \centering
    X = \{Q_1, Q_2, Q_3,….Q_n\}
\end{equation}

\begin{equation}
    \centering
    Z = \{A,B,C\}
\end{equation}

\begin{table}[]
\centering
\caption{VPR Techniques Employed in Each Conditional Variation Unit of the Switch-Fuse system .}
\begin{tabular}{|ccc|}
\hline
\multicolumn{3}{|c|}{\textbf{Conditional Variations}}                                                     \\ \hline
\multicolumn{1}{|c|}{\textbf{Seasonal}} & \multicolumn{1}{c|}{\textbf{Illumination}} & \textbf{Day-Night} \\ \hline
\multicolumn{1}{|c|}{AlexNet}           & \multicolumn{1}{c|}{HybridNet}                 & NetVLAD            \\ \hline
\multicolumn{1}{|c|}{AMOSNet}               & \multicolumn{1}{c|}{CoHOG}             & RegionVLAD         \\ \hline
\multicolumn{1}{|c|}{HOG}           & \multicolumn{1}{c|}{CALC}                  & HybridNet          \\ \hline
\end{tabular}
\end{table}

\subsection{ Applying Switching to Each Unit to Determine A Single Suited VPR Technique}
This step is divided into further sub-steps that helps determine a single technique as an output by each unit of the system. All of these sub-steps are performed individually for each component.

\subsubsection{Computing Probability that Query Image will be Correctly Matched (Posterior)}
This equation based on the Bayes theorem computes the posterior probability of the VPR technique correctly matching the image given the input query matching score. Where \textit{P(M)} is the probability of match by primary technique overall which is the prior probability of match which is obtained as discussed in detail in [13]. \textit{P(Z\textbar M)} is the likelihood for the VPR technique given it will correctly match for a certain matching score. This produces an updated but non-normalized probability distribution between the matching and mismatching. Finally, \textit{P(Z)} which is the marginalization in our equation is the summation of both updated non-normalized distribution of match and mismatch i.e  \textit{P(Z)} is the summation of \textit{P(Z\textbar M)*P(M)} and \textit{P(Z\textbar MM)*P(MM)}. Finally, \textit{P(M\textbar Z)} which is the calculated posterior, is used to determine the probability of correctly matching the given image. 

\begingroup
\small
\begin{equation}
\small
        {P(M|Z) = \frac{P(M)*P(Z|M)}{P(Z)}}
\end{equation}
\endgroup

\subsubsection{Determining VPR Technique for Switching}
Our posterior probability calculation allows us to predict the level of certainty or confidence with which the technique will correctly match the query image. While in case this value of probability is lower than our accepted value (0.5) the system attempts to switch to another technique complementary to the current primary technique. The system calculates the probability of complementarity that the primary technique has to the other available VPR techniques as descibred in step 3. Once the technique with the highest complementarity is determined the system switches towards this technique and determines the new posterior probability of matching the query image.

\subsubsection{Calculating Probabilities of Complementarity}
This equation computes the real time complementarity for the given query image that the primary technique has to the other available VPR methods in the system. Where \textit{P(Z\textsubscript{Q}\textbar M\textsubscript{A})} and \textit{P(Z\textsubscript{Q}\textbar MM\textsubscript{A})} is the probability of the certain score for query image given its matched or mismatched by technique A. While \textit{P(Z\textsubscript{Q}\textbar M\textsubscript{B})} and \textit{P(Z\textsubscript{Q}\textbar MM\textsubscript{B})} is the probability of the certain score event for \begin{figure} [tb]
    \centering
    \vspace*{1mm}
    \includegraphics[width=1\columnwidth]{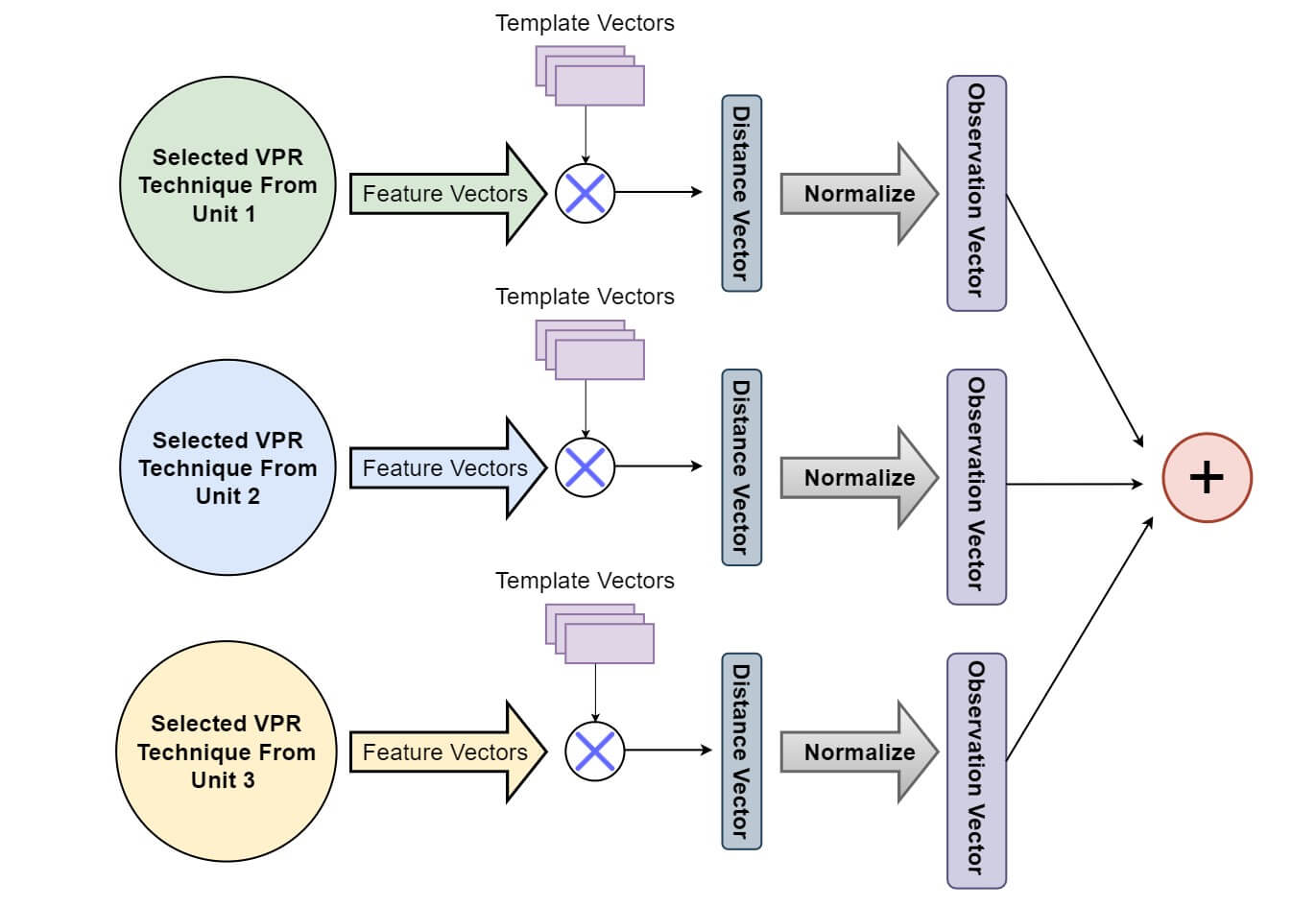}
    \caption{ The \textit{\textbf{Fusion}} step incorporates of how each selected VPR technique produces a distance vector which has the distance scores between the query image and reference images in database. Further we normalize these distance vectors prior to fusion which is performed by taking the summation of these normalized/observation vectors. }
    \label{fig:my_label}
\end{figure}
query image given its matched or mismatched by technique B. The equation computes the complementarity of A with B \textit{(CAB)} i.e the complementarity the two techniques, have to each other given a certain matching score i.e query image matching score. Finally, the system switches to the technique with the higher \textit{P(CAB)}.

\begingroup
\small
\begin{equation}
\small
        \textit{P(CAB)} = \frac{P(Z_Q|M_A)*P(Z_Q|M_B)}{P(Z_Q|MM_A)*P(Z_Q|MM_B)}
\end{equation}
\endgroup

\subsubsection{The Dynamic Switching}
The calculated posteriors for each technique are constantly checked against the threshold probability of above 0.5 to proceed. If the probability of match is below threshold the system will switch to another technique and perform the same steps to determine the probability of match until a suitable technique with satisfactory probability of match is found. In case no such technique can be found the system selects whichever technique has the highest probability of match. This selected technique is considered the output or prime selected technique of a single unit. 

\subsection{The Chosen VPR Technqiues}
The output of step two is a set of three selected VPR techniques, one selected by each unit from the tripartite model. Let N be the set of selected techniques for any given query image where $n \in N$ and n represents an individual selected VPR technique.

\begin{equation}
    \centering
    N = \{n_1,n_2,n_3\}
\end{equation}

\subsection{The Fusion}
The last and final step of the system is to fuse set N, the selected VPR techniques, to combine their results. Generally, to perform VPR each query image is compared against a database of prior images mainly by using different feature extractions and image matching techniques. This process results in a D dimensional similarity vector, which is a list of similarity scores between the query and reference images in the database. The similarity vector can be interpreted as the bigger the similarity score the stronger the chances of a correct match.

In the fusion step the selected VPR techniques with the highest probability of correctly matching the image are used simultaneously to fuse their similarity scores. In the currently tested system the number of selected techniques is limited to three but depending on computational resources and a different pool of techniques the number can be increased or decreased. Each n produces a similarity vector which has the similarity scores between the query image and reference images in database. Furthermore, as the set of techniques is arbitrary hence the distribution of these similarity scores within each technique may not be consistent with the distribution of other techniques. So, we normalize prior to fusing the set of N to ensure that each similarity vector has a minimum and maximum value of -0.001 and 0.999 respectively where $\epsilon=0.001$. (In case of normalized values falling under threshold a value of 0.001 is forcefully assigned.)

\begin{figure} [tb]
    \centering
    \vspace*{1mm}
    \includegraphics[width=1\columnwidth]{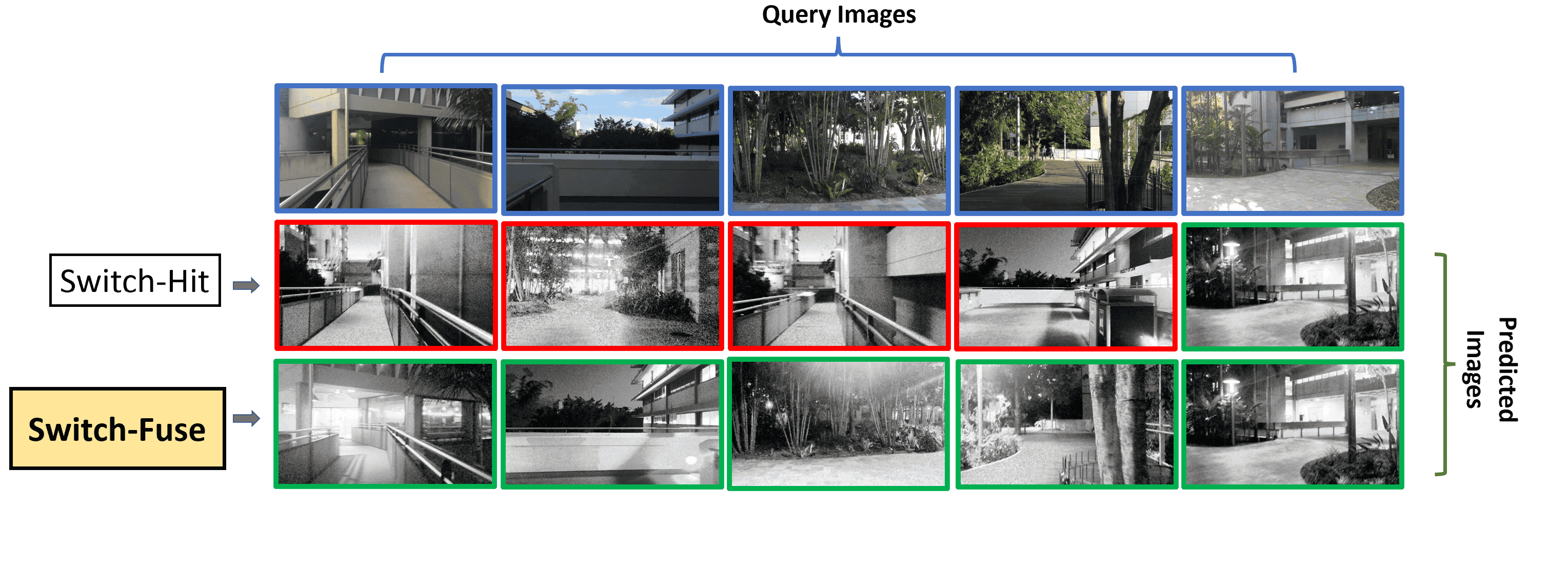}
    \caption{Different examples of query images using the SwitchHit [13] vs our proposed Switch-Fuse system to show how the latter outperforms over different queries. The images encapsulated by a blue frame present the queries while the red and green frame represent incorrect or correct matches respectively. In a series of example showcasing the better performance of the Switch-Fuse system an exceptional example of where both systems are successful is also added at the end.}
    \label{fig:my_label}
\end{figure}

\begingroup
\small
\begin{equation}
\small
        \hat{D}_{n}(i)=\frac{\hat{D}_{n}(i)-min({D}_{n})}{max({D}_{n})-min({D}_{n})}-\epsilon,  \forall_i  , n\in N
\end{equation}
\endgroup

The final step is to produce the combined similarity vector for fusion that is the $ D_{F} $. The matched image is the one with the maximum $ D_{F} $ score. 

\begingroup
\small
\begin{equation}
\small
{D}_{F}= \displaystyle\sum\limits_{n=1}^N \hat{D}_n
\end{equation}
\endgroup

\begin{figure*}[!htb]
    \vspace*{0.1in}
    \includegraphics[width=2\columnwidth]{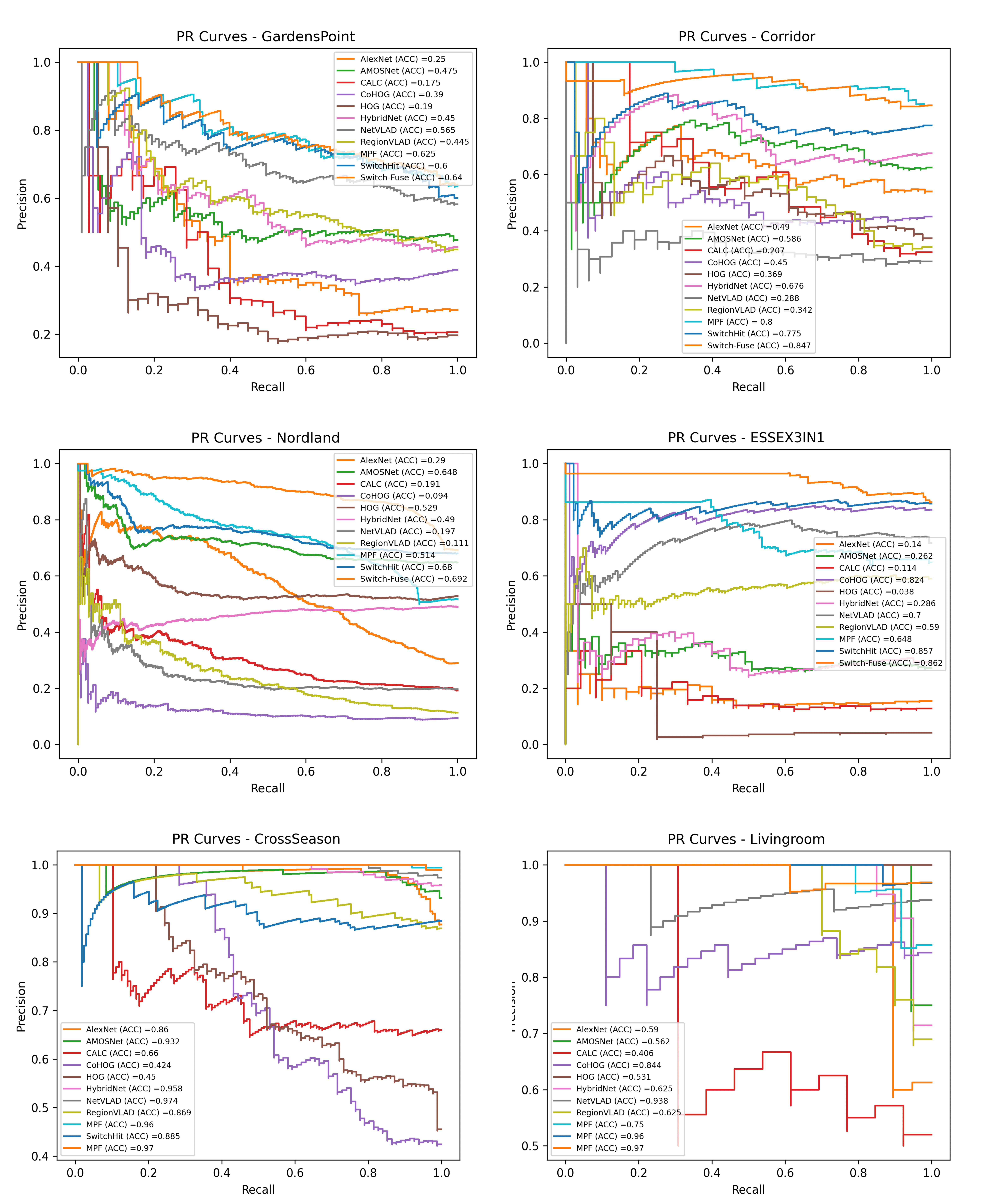}
   \caption{Precision-Recall curves showcasing performance of Switch-Fuse in comparison to  SwitchHit, MPF and other VPR methods on different data sets : GardensPoint (top left); Corridor (top right); Nordland (center left); Cross-Season (center right); ESSEX3IN1 (bottom left) and Livingroom (bottom right).}
   
    \label{figurelabel}
\end{figure*}

\section{Experimental Setup}
\begin{figure*}[!htb]
    \vspace*{0.1in}
    \includegraphics[width=17.5cm,height=7cm]{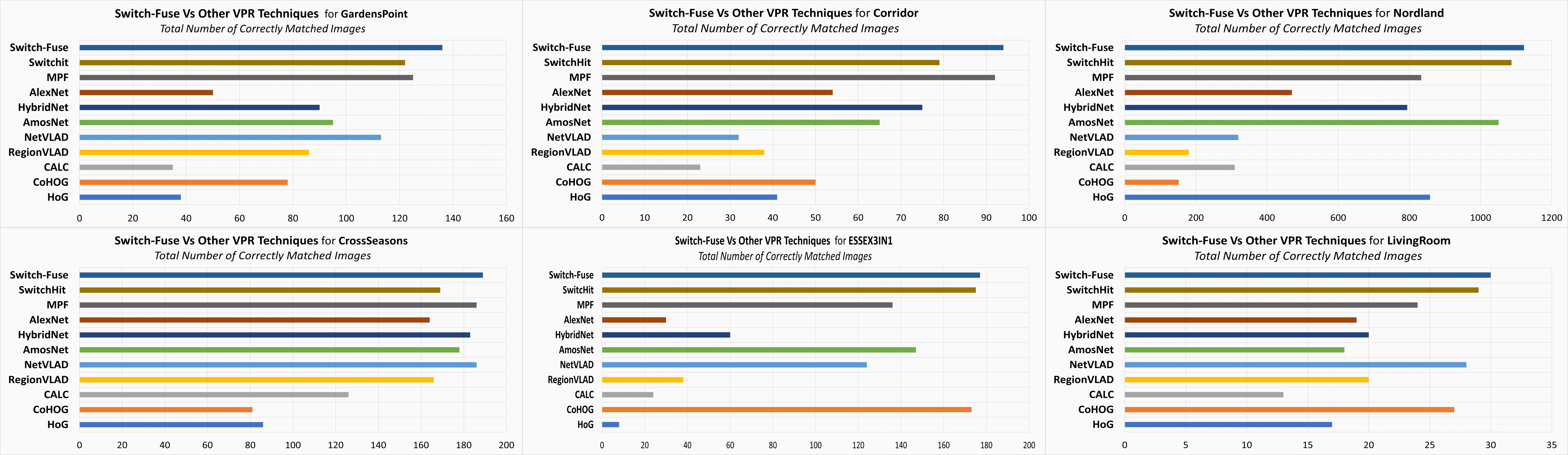}
   \caption{Performance improvement in terms of correctly matched images by Switch-Fuse in comparison to SwitchHit, MPF and other VPR methods on different data sets : GardensPoint (top left); Corridor (top center); Nordland (top left); Cross-Season (bottom left); ESSEX3IN1 (bottom center) and Livingroom (bottom right).}
    \label{figurelabel}
\end{figure*}

The Switch-Fuse system allows for the real-time selection of the most suitable VPR techniques for fusion given any query image, over different VPR data sets. TABLE I lists all the data sets along with their variations types including 
Corridor [38], Living room, ESSEX3IN1 [25], GardensPoint, Cross-Seasons [26], and Nordland [37]. We have selected a wide variety of datasets to ensure that each type of conditional variation is tested for the system. Table II presents the structure in which VPR techniques used have been employed in the proposed Switch-Fuse system. Each unit of different conditional variation is provided with three techniques that are theoretically known to be complementary pairs for the respective variation type [12]. The first unit consists of AMOSNet [20], HOG [30][34], and  AlexNet [28],[29]. The second unit employs CoHOG [35], HybridNet [20] and CALC [24]. Finally, the last unit consists of  HybridNet [20], NetVLAD [19],[22] and RegionVLAD [33]. This combination not only ensures the inclusion of complementary pairs but also all the state-of-the-art VPR techniques widely used for experiments. The implementation details of all these VPR techniques are the same as used in [39].

\section{Results}

In this section, we present the results employing the Switch-Fuse system to predict the best suited VPR techniques for fusion while the performance is presented via accuracy and PR-curves. Additionally we also present results on an image-by-image basis to show the increase in total correct matches over widely employed VPR data sets. The improvement produced in the results is accounted by accurate prediction and selection of the best suited VPR techniques and then their fusion, as further explained via different examples.  

Beginning from Fig. 5 which compares our Switch-Fuse system to the SwitchHit [13] system, Multi-Process Fusion (fusing three best performing VPR techniques overall) [10] and as well as other VPR techniques to showcase the difference in performance using PR curves. The results and improvements vary over different data sets due to their extremely varying environments and sizes. For example testing out the GardensPoint where we observe an accuracy of 0.64 over the data set and this is an significant improvement over SwitchHit, MPF or any other VPR techniques in comparison. Similarly, for the Corridor data set Switch-Fuse produces an overall accuracy of 0.84 which again is not only higher in comparison to all single VPR techniques but both SwitchHit and MPF as well. Switch-Fuse is able to achieve an accuracy of almost 0.7 for Nordland data set which again is significantly higher than both SwitchHit and MPF. It is also a good example to observe how in some cases SwitchHit outperforms MPF and vice versa but together they outperform each other in all cases.  
Here it would be correct to conclude that there is a trade-off between the level of performance and only fusing chosen techniques by switching rather than random multiple VPR techniques. Empirical data for different techniques that should be suitable together for fusion is not always true for all data sets or query images and the Switch-Fuse system as evident by the results helps predict VPR techniques in real-time which are actually useful to be fused. Similar results for other data sets including CrossSeasons, ESSEX3IN1 and Livingroom are presented depicting the higher accuracy Switch-Fuse was able to achieve in comparison to SwitchHit or MPF on these data sets.  After testing Switch-Fuse for a series of varied data sets we conclude that the system is in fact able to boost accuracy performance over different environmental variations by performing informed switching, and then fusing these selected techniques only.

Fig. 6 depicts a comparative analysis between overall performance on each data set tested on the basis of total increase in number of correctly matched images. It is a simple way to observe over an image-by-image basis how the overall accuracy of the system is better than other systems in comparison. The first data sets presented are the GardensPoint and Corridor data set where an improvement of 15 or more images than SwitchHit [13], MPF [10] and even significantly more images than other individual VPR techniques can be observed. The Switch-Fuse system has around 60 more correctly matched images for the Nordland data set than any other option available throughout experimentation. For the CrossSeason and ESSEX3IN1 data sets an improvement of 3 to 4 images can be observed which is still higher than even the best performing VPR technique, MPF or SwitchHit [13]. ESSEX3IN1 is one of the examples to show the capability of Switch-Fuse where MPF has lower performance than a single technique (CoHOG) which is trained specifically for the data set, while SwitchHit has very minor improvement but together they outperform any options available including CoHOG. A pattern of significant improvement in performance by the Switch-Fuse system helps conclude that the hybrid model of Switch-Fuse allows for an intelligent real time switching to the most suitable VPR techniques to be fused and the fusion methodology further boosts performance. 

Furthermore, Fig. 4 and Fig. 7 are actual representations of examples taken from our experiment to show case how the Switch-Fuse system performs. Fig. 7 explains this over the GardensPoint data set to show the different VPR techniques selected, on each query image, to be fused. It is important to mention that although mostly a combination of the same techniques can be observed over a data set this does change, more in some cases than the others. Although it is possible for each query image to have its own combination of VPR techniques for fusion, a certain level of uniformity can be observed over the data set with mostly the same combination being selected. Furthermore an example illustrating a case where none of the three individual VPR techniques are able to correctly match the query but the selection of the three specific technique and their fusion results in a successful match. Many similar cases can be observed overall on multiple queries that result in the performance improvement seen using Switch-Fuse.



\section{Conclusions and Discussion}
In this work we propose a hybrid system designed to incorporate the properties of both, dynamic switching and fusion systems. Our proposed method not only attempts to overcome the shortcomings that both systems posses individually but forms an amalgamation of the strengths of these two otherwise discrete approaches. The Switch-Fuse system aims to select the best suited VPR technique from a series of well curated three tiers on the basis of complementarity and environmental variations. While the selected VPR techniques undergo the fusion process and produce a combined similarity vector that is used to correctly match the query image. The results demonstrate a significant improvement as evident by the increase in overall performance accuracy that advocate that the proposed system was successful in performing well over a variety of environmental variations. This work paves path for many other innovative and unconventional ideas that can be used for improvement of VPR performance. An important observation to be made is that basis of switching and selection of different VPR techniques in this work is complementarity, however, a different metric for switching and selection can be interesting work for the future. 
\begin{figure} [tb]
    \centering
    \vspace*{1mm}
    \includegraphics[width=1\columnwidth]{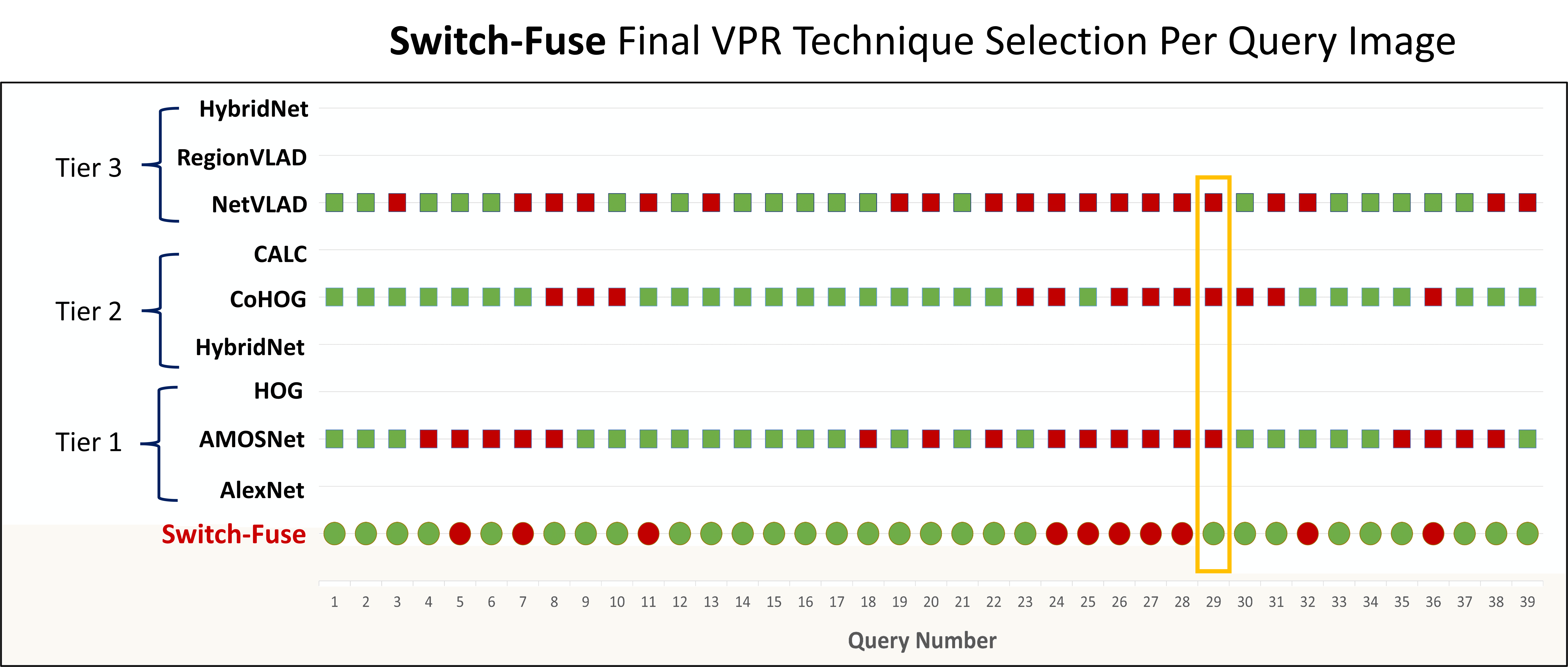}
    \caption{Presents the Switch-Fuse Final Technique selections per query for the GardensPoint data set as an example. The green and red blocks represent a match or mismatch by any individual VPR technique while the green and red circles represent match or mismatch by the Switch-Fuse system. An example of how the careful selection via switching and fusing results in correct matches is presented by the yellow window. The example presents a case where none of the techniques individually are able to correctly match the query but Switch-Fuse results in a successful match. }
    \label{fig:my_label}
\end{figure}


\end{document}